\documentclass{article}
\usepackage{spconf,amsmath,epsfig}
\usepackage{times}
\usepackage{epsfig}
\usepackage{graphicx}
\usepackage{amsmath}
\usepackage{amssymb}
\usepackage{algorithmic}
\usepackage{algorithm}
\usepackage{indentfirst}
\usepackage{array}
\usepackage{subfig}
\usepackage{multirow}
\usepackage[pagebackref=true,breaklinks=true,letterpaper=true,colorlinks,bookmarks=false]{hyperref}

\begin{document}
\topmargin=0mm

\def\x{{\mathbf x}}
\def\L{{\cal L}}

\title{Local- and Holistic- Structure Preserving Image Super Resolution via Deep Joint Component Learning}
%
%
\name{Yukai Shi$^1$, Keze Wang$^1$, Li Xu$^2$, Liang Lin$^{1\star}$ \thanks{Corresponding author is Liang Lin.}}
\address{$^1$Sun Yat-Sen University, Guangzhou, China \\
$^2$SenseTime Group Limited \\
\small {shiyk3@mail2.sysu.edu.cn, kezewang@gmail.com, xuli@sensetime.com, lianglin@ieee.org}}

\maketitle
\begin{abstract}
Recently, machine learning based single image super resolution (SR) approaches focus on jointly learning representations for high-resolution (HR) and low-resolution (LR) image patch pairs to improve the quality of the super-resolved images. However, due to treat all image pixels equally without considering the salient structures, these approaches usually fail to produce visual pleasant images with sharp edges and fine details. To address this issue, in this work we present a new novel SR approach, which replaces the main building blocks of the classical interpolation pipeline by a flexible, content-adaptive deep neural networks. In particular, two well-designed structure-aware components, respectively capturing local- and holistic- image contents, are naturally incorporated into the fully-convolutional representation learning to enhance the image sharpness and naturalness. Extensively evaluations on several standard benchmarks (\emph{e.g.}, \textit{Set5}, \textit{Set14} and \textit{BSD200}) demonstrate that our approach can achieve superior results, especially on the  image with salient structures, over many existing state-of-the-art SR methods under both quantitative and qualitative measures.

\end{abstract}

\begin{figure*}[ht]
\centering
\includegraphics[width=1\textwidth]{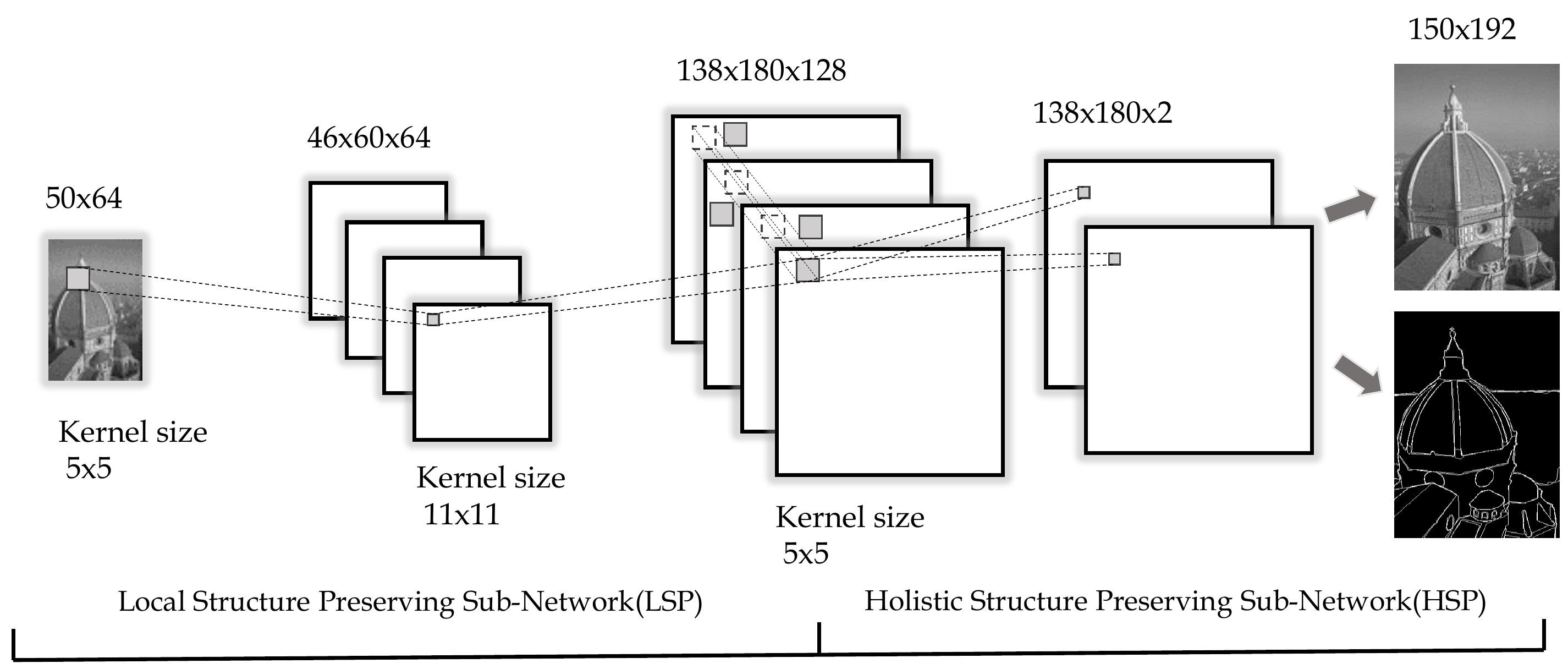}
\caption{The architecture of the proposed local- and holistic- structure preserving neural networks. The neural networks are stacked by a convolutional layer, a deconvolution layer, pixel placement operation and a convolution layer. The first two layers and the operation forms the local structure preserving sub-network (LSP), while the final convolutional forms the holistic structure preserving sub-network (HSP). The LSP upsamples the LR patches via deconvolution with local structure preserving displacement (Pixel Placement). Then the HSP further refines the output of LSP via convolution with a multi-task of holistic structure preserving objective. }
\label{fig:overall-pipeline}
\vspace{-15pt}
\end{figure*}

\section{Introduction}
\label{sec:intro}
Owing to various practical reasons such as cost of camera, storage limitation and limited bandwidth, images with low resolution are inevitable. To solve this problem, single image super resolution (SR), with the goal of increasing the resolution of the image from the single input, has been drew considerable attention from different research communities.

Techniques for single image SR can be roughly categorized as reconstruction-, example- and interpolation- based approaches. The reconstruction-based approaches \cite{irani1991improving} assume that the registered frames are in compliance with a global blur degradation model, which can be formulated by explicitly modeling deconvolution \cite{shan2008fast} or by allowing blind super resolution with unknown blur degradation \cite{michaeli2013nonparametric}. Due to the inverse problem with inaccurate blur kernels, reconstruction-based approaches may introduce ringing artifacts around salient structures \cite{michaeli2013nonparametric}.

Opening up a new data-driven direction for single image SR, example-based approaches \cite{glasner2009super,kim2010kk, yang2010sc, freedman2011image} advance in the use of internal or external patch data to increase the resolution by a large factor with synthesized artificial details. However, the synthesized details are not necessarily consistent with real details. Specifically, it is not well grounded that the synthesized patch does bring real details of original optical images into the low resolution version.

Though having the advantage in efficiency and balanced performance,  interpolation-based approaches, such as bilinear / bicubic / spline, are prone to mix colors along the main edges, especially when encountering large upsampling factors. Most traditional interpolation-based approaches, which use single fixed kernel to interpolate the whole image, are inevitable to blur the main structure, e.g., \cite{chu2008gradient,van2012polygon} propose a kind of adaptive-interpolation and attempt to resolve this problem. The pursuit of bringing higher visual quality to human visual system is still the fundamental goal of many practical SR approaches. As a matter of fact, human visual system is more sensitive to color and structural changes than absolute color values. Preserving the contrast and sharpness of the salient structural edge is thus crucial for generating visually plausible results. In contrast, some SR approaches \cite{NIPS2015xuli,dong2014srcnn} take interpolation operation first. This may bring negative influences on the main structure because of not good enough initialization. It is beneficial if we can incorporate some techniques to learn multiply edge-preserving kernels for interpolation from sufficient image data.

Deep convolutional neural network has been successfully applied to various computer vision tasks including rain/dirt removal \cite{eigen2013restoring}, image deconvolution \cite{xu2014deep}, noise removal \cite{jain2009natural} and image in-painting \cite{xie2012image}. More recently, \cite{wang2015deep,NIPS2015xuli,dong2014srcnn, cui2014deep, Zhang2015Joint} have achieved promising results by incorporating deep learning for SR. In particular, example-based approaches with deep learning framework~\cite{cui2014deep,dong2014srcnn} propose to capture the mapping between low- and high- resolution images to obtain the state-of-the-art performance. Cui \emph{et al.}\cite{cui2014deep} advocated the use of cascade networks for super resolution, with a local auto-encoder architecture. Dong \emph{et al.} \cite{dong2014srcnn} applied the similar FCN in super resolution. However, these approaches are less optimal for that they tend to interpolate image with pixels equally treated and actual contrast of border/texture ignored during the training procedure.

In this paper, we present a local- and holistic- structure preserving neural networks, aiming for salient structural edge preservation. As illustrated in Figure~\ref{fig:overall-pipeline}, the proposed model is stacked by two component named Local Structure Preserving sub-network (LSP) and Holistic Structure Preserving sub-network (HSP). The LSP  upsamples the input low resolution patches via a deconvolution network with local structure preserving displacement (Pixel Placement). The HSP further refines the output of LSP via a fully convolutional layer with a multi-task of structure preserving objective.

The \textbf{main contributions} of this paper are organized as follows: (1) We propose a novel deep neural network for super-resolution, which achieves state-of-the-art performance with a small architecture on public \textit{Set5} \cite{bevilacqua2012low}, \textit{Set14} \cite{zeyde2012single} and \textit{BSD200} \cite{amfm_pami2011} benchmarks. (2) We make an attempt to re-design the classical interpolation pipeline by replacing building blocks with content-adaptive and structure-aware neural networks. (3) Our study demonstrates that the proposed joint local- and holistic- structure preserving can significantly benefit super resolution.

\section{Framework}
Figure ~\ref{fig:overall-pipeline} illustrates the architecture of our proposed local- and holistic- structure preserving framework, which consists of two components called Local Structure Preserving sub-network (LSP) and Holistic Structure Preserving sub-network (HSP). The LSP is stacked by a convolutional layer, a deconvolutional layer and our proposed pixel placement operation in order to preserve the local structure of the image. Focusing on the holistic structure of the image, the HSP employs one convolutional layer to further refine the result of LSP by considering non-local and boundary information.


\subsection{Local Structure Preserving sub-network (LSP)}
Different from existing SR approaches that typically adopting edge-blurring interpolation techniques in the initialization step~\cite{dong2014srcnn, glasner2009super}, our proposed LSP  incorporates local deconvolution and structure preserving pixel placement into the interpolation phase.

\begin{figure}[t] \centering
\centering
\includegraphics[width=0.6 \columnwidth]{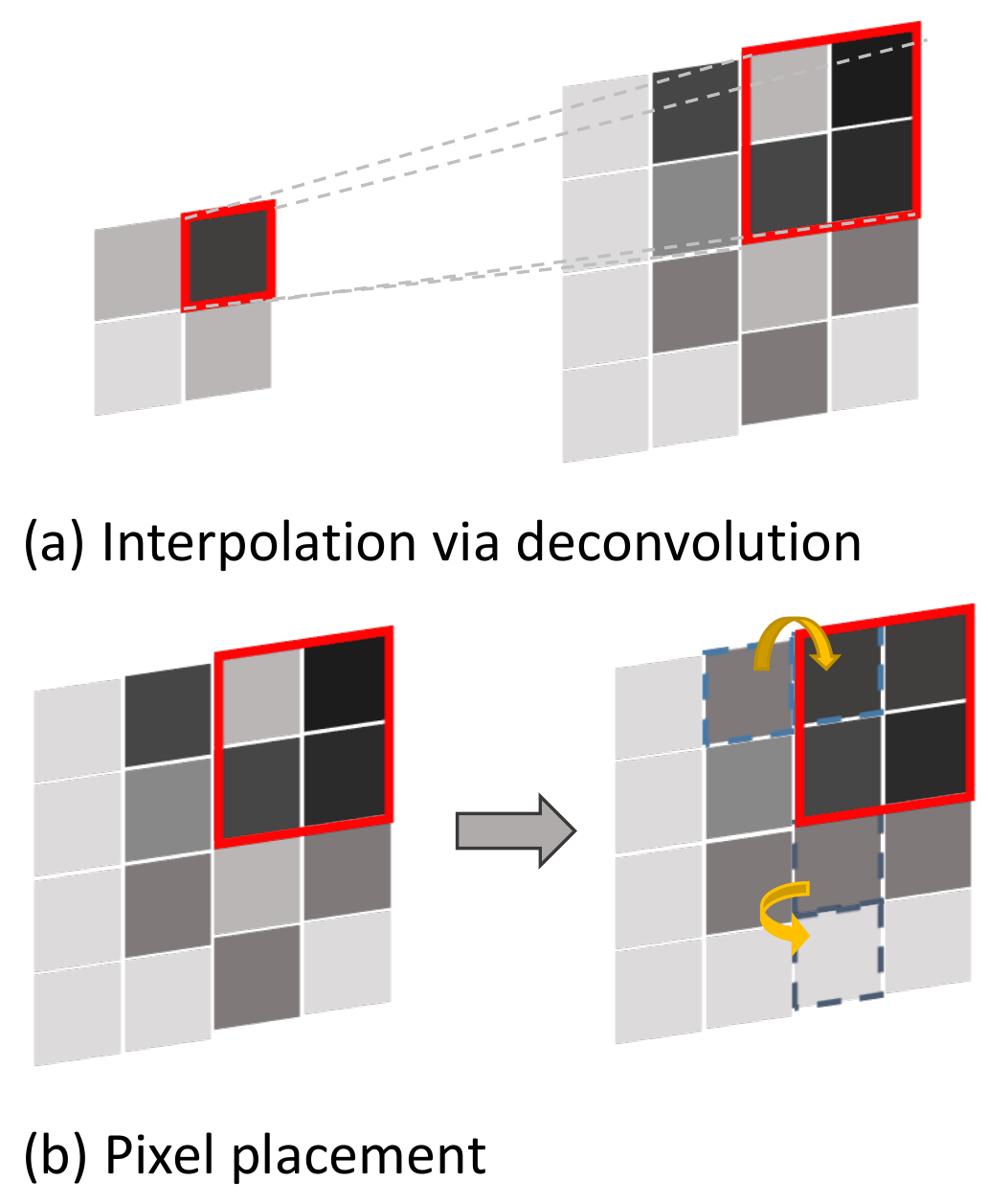}
\vspace{-8pt}
\caption{Illustration of the processing of interpolation by deconvolution. }
\label{fig:LSP-figure1}
\vspace{-18pt}
\end{figure}

\emph{\textbf{Deconvolution.}}~Traditional interpolation assumes that the LR pixels are evenly placed in the HR grid. In this regard, interpolation is a linear transform invariant (LTI) operation. Specifically, denote the intensity of each pixel in LR and HR image by ${x_i}$ and $y_j$, respectively. The interpolation, e.g., bilinear, can be expressed as
\begin{equation}
y_j = \sum_{n_i \in \Omega}{\omega_{n_i} \cdot x_{n_i}},
\end{equation}
where $\Omega$ is the local window, $n_i$ indexes the pixels and $\omega_{n_i}$ indicates the fixed bilinear weights.
As one can see that, the formulation to calculate $y_j$ is similar to the deconvolution operation, which is defined as
\begin{equation}
h^{l} = \sigma ( W_1 * h_1^{l-1} + \cdots  W_n * h_n^{l-1} + b ),
\end{equation}
where $[W_1, W_2, ..., W_n]$ indicates the kernel weight mapping from $(l-1)^{th}$ to the $l^{th}$, $b$ is the bias. $h^{l}$ and $h^{l-1}$ mean output and input, respectively. $\sigma(\cdot)$ is the nonlinear function. This process is illustrated in Figure~\ref{fig:LSP-figure1}{\color{red}a}. Hence, the interpolation can be accomplished via a deconvolution layer. Moreover, the deconvolution kernel can be adaptively learned from sufficient training data.

\begin{figure} [t]\centering
\subfloat[][ \centering Bicubic kernel \par PSNR 32.71] {
\includegraphics[width=0.42\columnwidth] {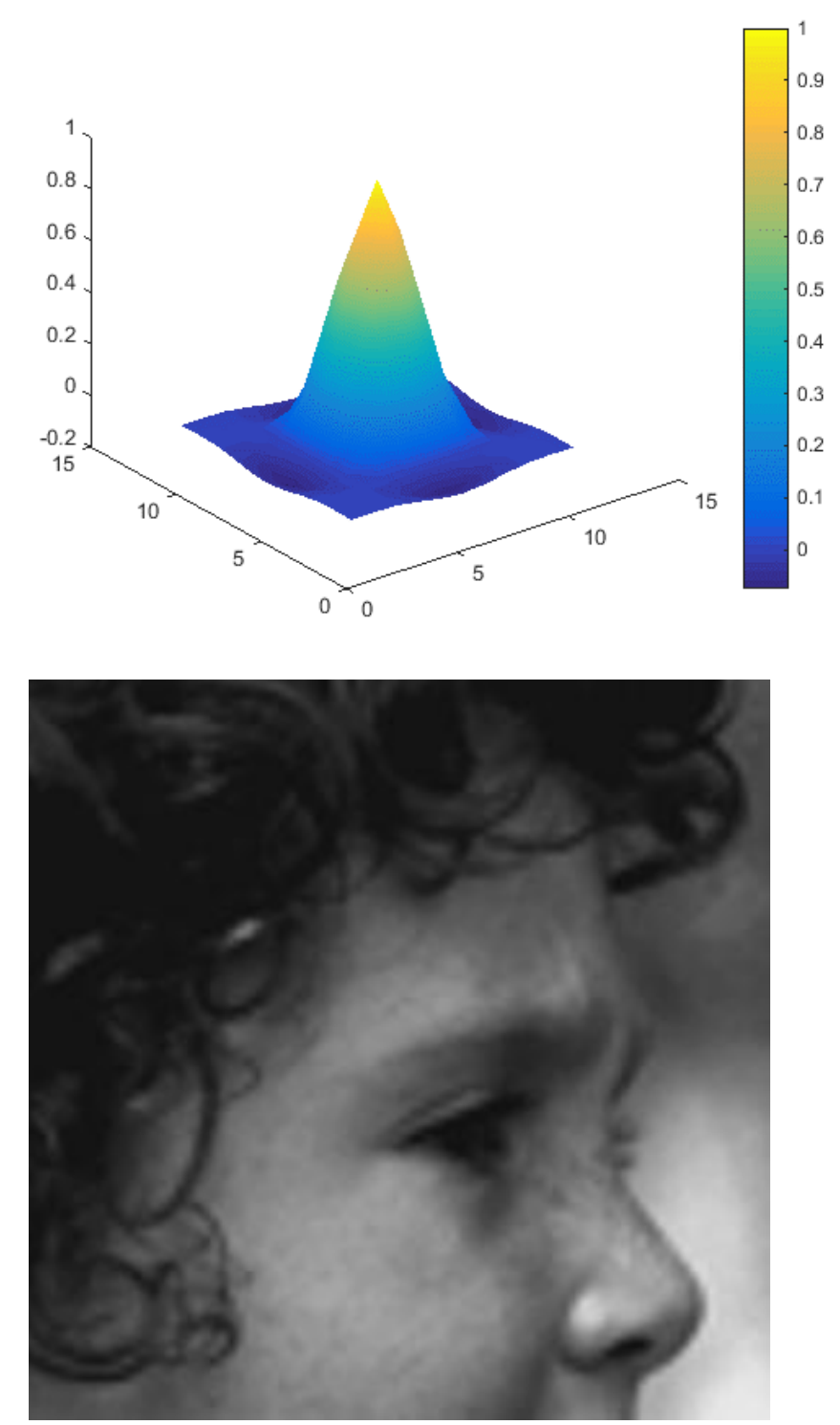}
}
\subfloat[][ \centering Learned  kernel \par  PSNR 33.10] {
\includegraphics[width=0.42\columnwidth]{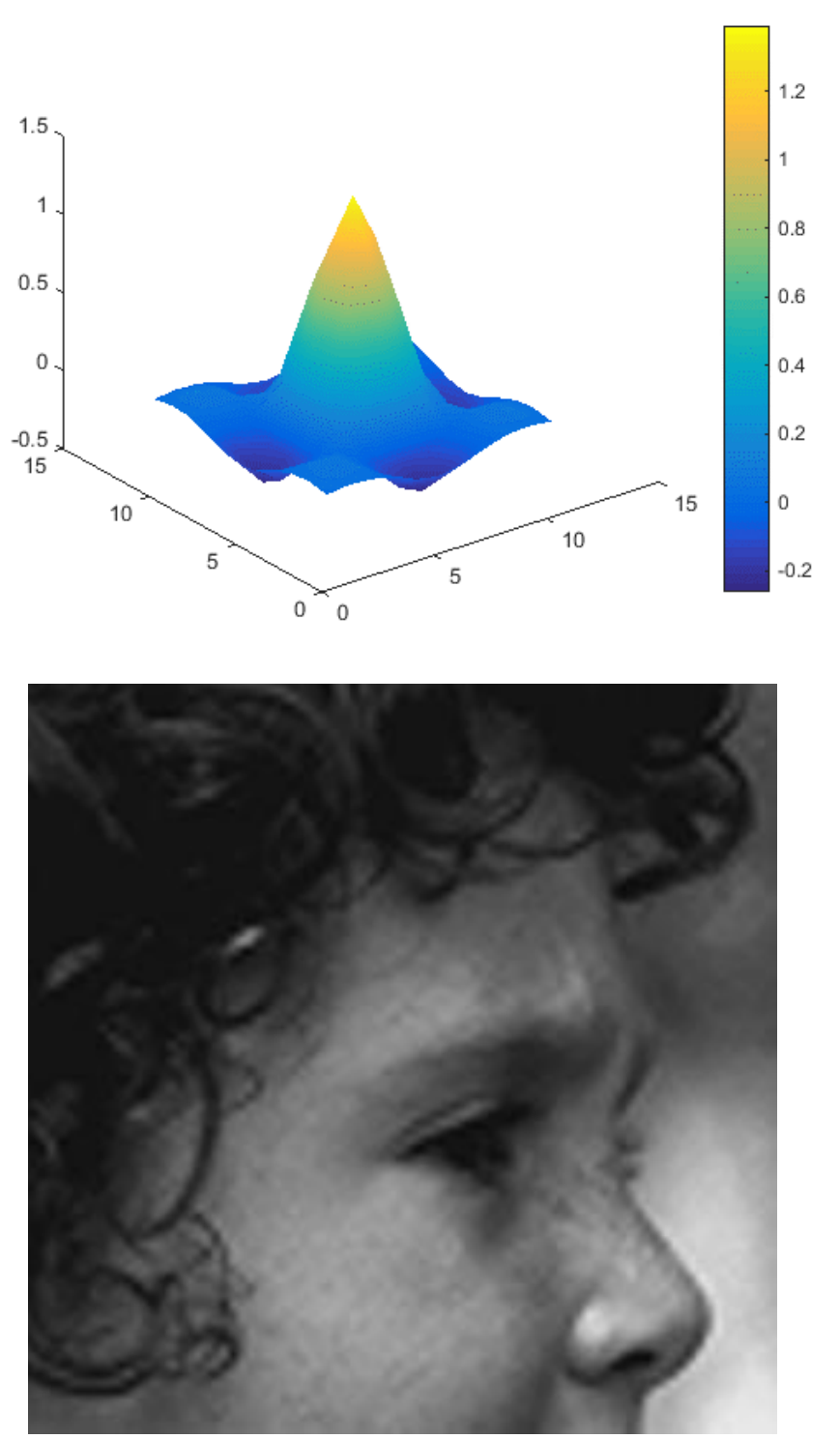}
}
\vspace{-8pt}
\caption{Results of the bicubic and learned interpolation operators. }
\label{fig:LSP-figure2}
\vspace{-15pt}
\end{figure}

We propose to learn a local deconvolution kernel from sufficient image pairs for interpolation. Figure~\ref{fig:LSP-figure2} visualizes some motivating results via the learned kernels from corresponding images. Compared with bicubic, the result reveals that the learned kernel is more extraordinary and its performance is superior to the traditional fast interpolation approach.

\begin{table*}[tb]  \centering \footnotesize 
\center
\begin{tabular}{|c|*{14}{>{\hfil}p{23pt}<{\hfil}|}}
\hline
Test set & Scale & \multicolumn{2}{|c|}{Bicubic} & \multicolumn{2}{|c|}{NELLE \cite{chang2004nelle}} & \multicolumn{2}{|c|}{SCIP \cite{kim2010kk}}  & \multicolumn{2}{|c|}{ANR \cite{timofte2013anr}}    & \multicolumn{2}{|c|}{SRCNN \cite{dong2014srcnn}}  & \multicolumn{2}{|c|}{Ours} \\
\hline
 &  & PSNR & SSIM & PSNR & SSIM & PSNR & SSIM & PSNR & SSIM & PSNR & SSIM & PSNR & SSIM  \\
 \hline
 \hline
 \multirow{3}{*}{Set5} & 2 & 33.66 & 0.9299 & 35.77 & 0.9489 & 32.20 & 0.9511 & 35.83 & 0.9499 & 36.43 & 0.9515 & \textbf{36.90} & \textbf{0.9547} \\
   & 3 & 30.39 & 0.8677 & 31.84 & 0.8946 & 32.28 & 0.9033 & 31.92 & 0.8958  & 32.56 & 0.9049& \textbf{32.79} & \textbf{0.9090} \\
   & 4 & 28.42 & 0.8099 & 29.61 & 0.8391 & 30.03 & 0.8541 & 29.69 & 0.8408  & 30.31 & 0.8587& \textbf{30.51} & \textbf{0.8636} \\
   \hline
 \multirow{3}{*}{Set14} & 2 & 30.23 & 0.8689 & 31.76 & 0.8992 & 32.11 & 0.9026 & 31.80 & 0.9004 & 32.39 & 0.9042  & \textbf{32.68} & \textbf{0.9079} \\
   & 3 & 27.54 & 0.7742 & 28.60 & 0.8080 & 28.94 & 0.8132 & 28.65 & 0.8096 &  29.13 & 0.8163 &  \textbf{29.31} & \textbf{0.8208} \\
   & 4 & 26.00 & 0.7026 & 26.81 & 0.7334 & 27.14 & 0.7419 & 26.85 & 0.7355 & 27.40 & 0.7486 & \textbf{27.51} & \textbf{0.7523} \\
   \hline
    \multirow{3}{*}{BSD200} & 2 & 29.43 & 0.8538 & 30.57 & 0.8879 & 31.23 & 0.8997 & 30.61 & 0.8886  & 31.49 & 0.9055  & \textbf{31.78} & \textbf{0.9071} \\
   & 3 & 27.18 & 0.7621 & 27.89 & 0.7948 & 28.13 & 0.8014 & 27.92 & 0.7962 & 28.33 & 0.8093  & \textbf{28.47} & \textbf{0.8112} \\
   & 4 & 25.92 & 0.6955 & 26.47 & 0.7240 & 26.63 & 0.7284 & 26.50 & 0.7258  & 26.84 & 0.7384  & \textbf{26.94} & \textbf{0.7408} \\
\hline

\end{tabular}
\vspace{-8pt}
\caption{Comparison between our models and other methods on the PSNR/SSIM indexes. We use the \textbf{bold face} to label the first place in each track.}
\label{table:PSNR/SSIM-on-different-method}
\vspace{-10pt}
\end{table*}

\begin{figure} [t]\centering
\subfloat[One dimension HR sample] {
\includegraphics[width=0.9\columnwidth]{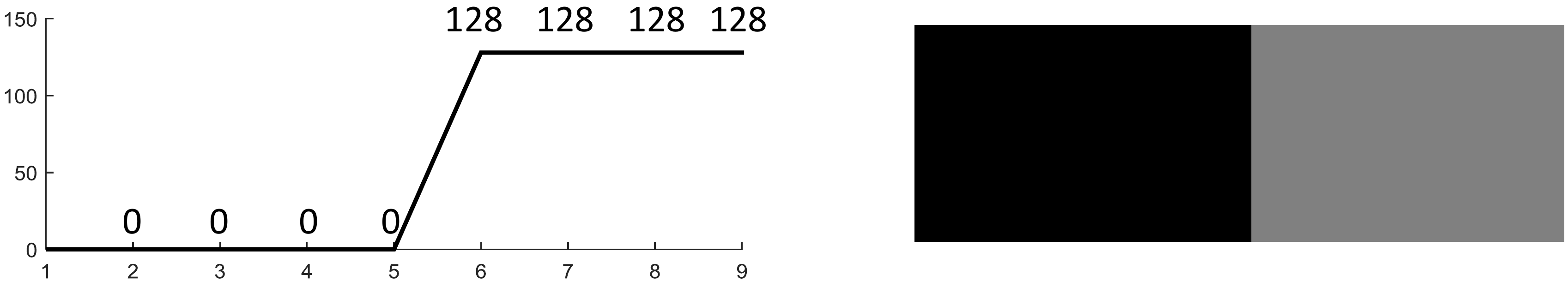}
\label{fig:comparison-on-HSP-and-noHSP-a}
}
\\
\subfloat[Bilinear kernel without LSP-displacement] { \label{fig:comparison-on-HSP-and-noHSP-b}
\includegraphics[width=0.9\columnwidth]{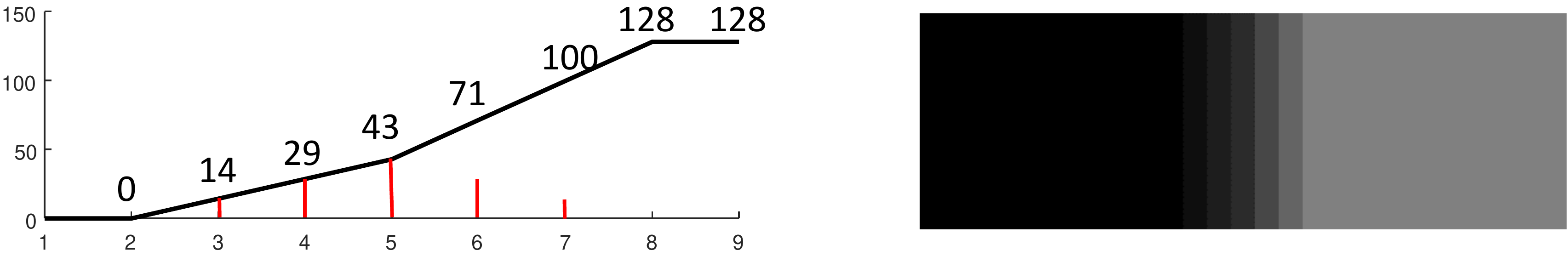}
}
\\
\subfloat[Bilinear kernel with LSP-displacement] { \label{fig:comparison-on-HSP-and-noHSP-c}
\includegraphics[width=0.9\columnwidth]{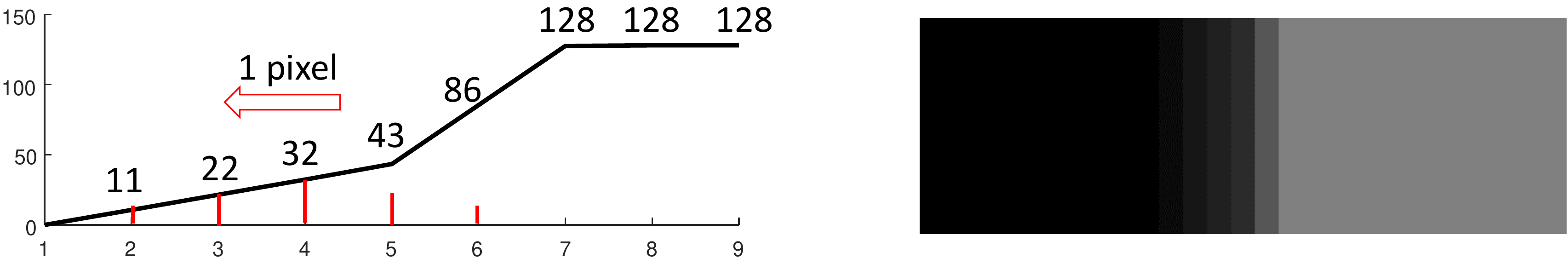}
}
\vspace{-10pt}
\caption{Illustration of the the pixel placement problem. (a) the original HR sample; (b) the result of bilinear upsampling; (c) By moving LR pixel away from the edge in HR grid, the interpolated edge becomes much clearer. }
\label{fig:LSP-displacement}
\vspace{-15pt}
\end{figure}

\emph{\textbf{Pixel Placement.}}~ As mentioned above, the single local deconvolution for interpolation is a linear translation invariant (LTI) operator. Therefore, when applying the LTI filter to pixels with large contrast in LR image, the color mixing is unavoidable because of the actual interaction range between pixels, i.e., image edges are blurred in this case. To overcome this problem, we propose an edge-preserving method called Pixel Placement, which slightly moves the original position of LR pixels in its HR grid to make HR grid homogeneous (see Figure~\ref{fig:LSP-figure1}{\color{red}b} for more details). Figure \ref{fig:comparison-on-HSP-and-noHSP-a} illustrates a LR sample from the original HR image. Comparing with the result from pure bilinear interpolation in Figure \ref{fig:comparison-on-HSP-and-noHSP-b}, refining LR pixels in the HR grid (e.g., moving LR pixels farther from the structural edge) contributes to the sharper edge in HR (see Figure \ref{fig:comparison-on-HSP-and-noHSP-c}).

The local structure preserving interpolation is performed by combining above mentioned deconvolution and pixel-placement inside the neural networks. It should be noted that when the pixel is not evenly placed, the interpolation can not be approximated by one deconvolution. The unevenly distributed pixels may have different influence on pixels, which makes the operation no longer LTI. The shepard interpolation theory \cite{shepard1968two} addressed the problem by normalizing via an indicator map with value one on those unevenly placed sparse LR pixels. The deconvolution is applied both on the HR grid with LR pixels and the indicator map. The normalization
is conducted by element-wise division as:
\vspace{-4pt}
\begin{equation}
h_{LSP}(x_i) = \frac{d(f_l( x_i ))}{f_l(\mathbf{1}(x_i))}  ;
\end{equation}
where $x_i$ denotes the input LR image, $f_l(\cdot)$ denotes the deconvolution network, $d(\cdot)$ denotes the image grid guided by the proposed pixel placement and $\mathbf{1}(x_i)$ denotes the indicate map with value on those unevenly placed sparse pixels of $x_i$.


\subsection{Holistic Structure Preserving Sub-network (HSP)}

The LSP stands by itself as a single image super resolution network. We aim to improve the performance of SR on the main structural edges, with the help of human labeled boundary guidance. Structure edges and contours are of great importance for human vision system, although they are not necessarily associated with good quantitative results. The HSP first takes the LSP output $h_{LSP}$ as input, i.e., LSP and HSP are combined together in such an end-to-end fine-tuning manner. Moreover, we additionally add an auxiliary objective for global boundary predictions, which guides the network to obtain the ability of preserving the global structure. Specifically, we train our model on a set of (LR, HR, boundary) triplets. In general, suppose we generate $N$ samples in the training set with each sample containing one LR image $x_i$ , one groundtruth HR image $y_i$, and one boundary image $b_{i}$. Denote the parameters of LSP and HSP by $\omega_l$ and $\omega_h$, respectively, our model can be formulated as:
\begin{equation}
\label{equ:goal}
\min_{\omega_l, \omega_h} \frac{1}{N} \sum_{i=1}^N \big( [y_i, \alpha \cdot b_i] - f_h(h_{LSP}(x_i; \omega_l); \omega_h ) \big)^2
\end{equation}
where $f_h(\cdot)$ denotes two outputs (i.e., 2 feature maps) of the proposed HSP with the multi-task objective. $[y_i, \alpha \cdot b_i]$ denotes the fitting targets and $\alpha$ controls weight of preserving global structure.

\subsection{Model Training and Testing}
As our proposed model seamlessly integrates local structure preserving and global structure preserving, the standard back propagation algorithm is applicable to optimize the model parameters \{$\omega_l, \omega_h$\}. The only remark is that the pixel placement operation is non-differentable, thus we need to save the original location of the moved pixels for the back propagating from HSP to LSP.  As for testing the image $x_i$, the predict result is inside $f_h(h_{LSP}(x_i; \omega_l); \omega_h )$.

\vspace{-5pt}
\section{Experiments}

\setlength{\abovecaptionskip}{2pt}
\setlength{\belowcaptionskip}{-3pt}
\begin{figure*}[htbp]\centering
\subfloat[][ \centering NELLE \cite{chang2004nelle} \par ]  {
\includegraphics[width=0.37\columnwidth]{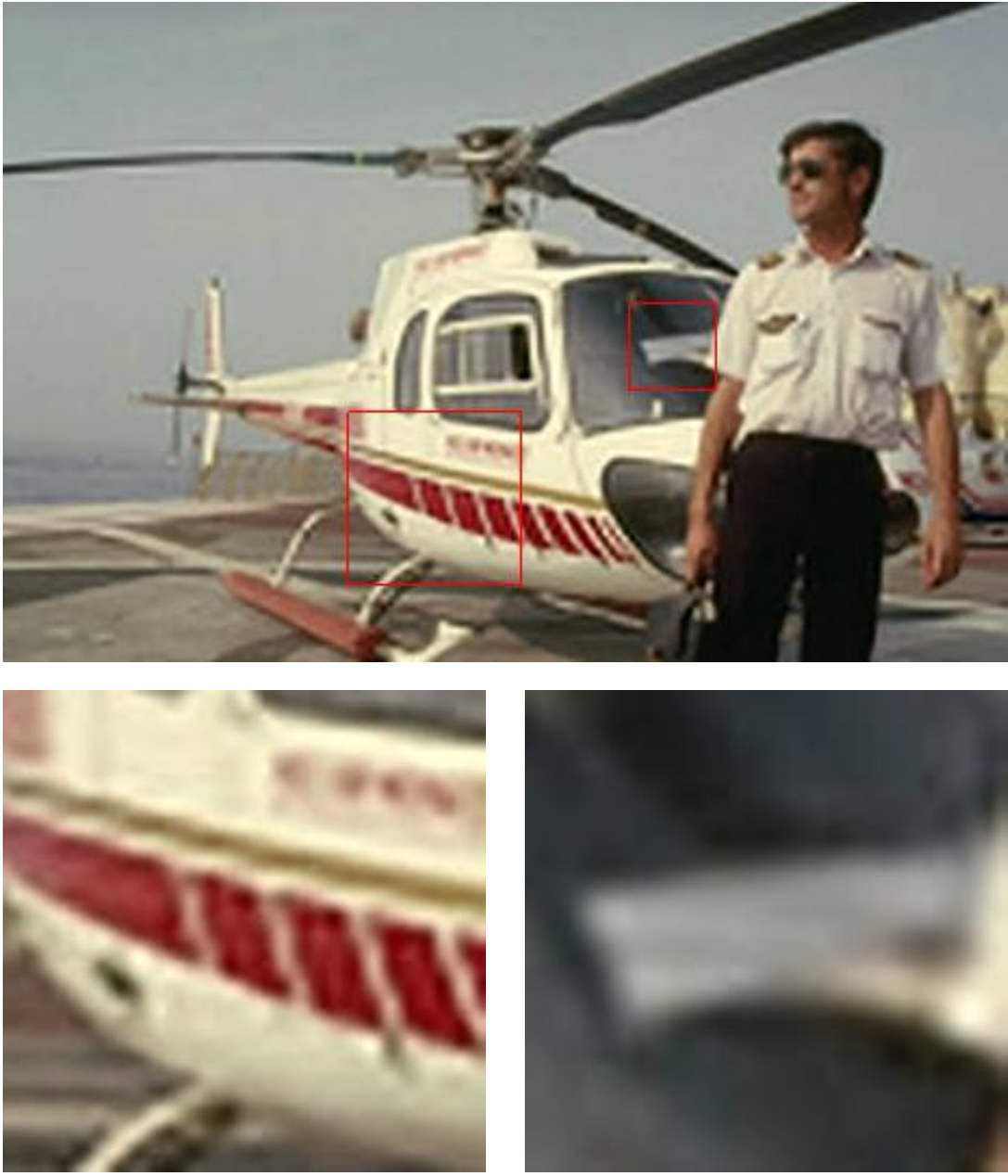}
}
\subfloat[][ \centering SCIP \cite{yang2010sc} \par ]  {
\includegraphics[width=0.37\columnwidth]{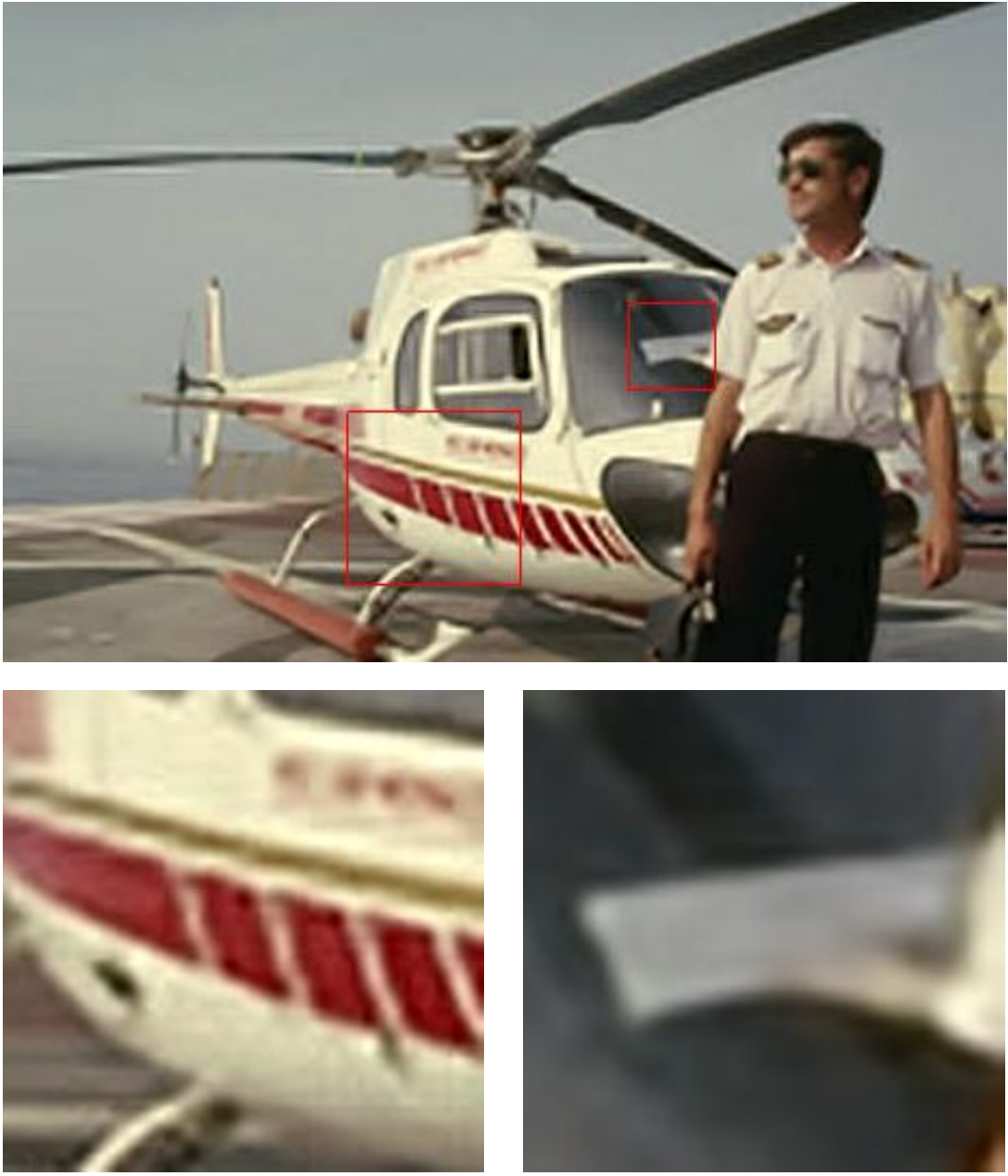}
}
\subfloat[][ \centering ANR \cite{timofte2013anr} \par]  {
\includegraphics[width=0.37\columnwidth]{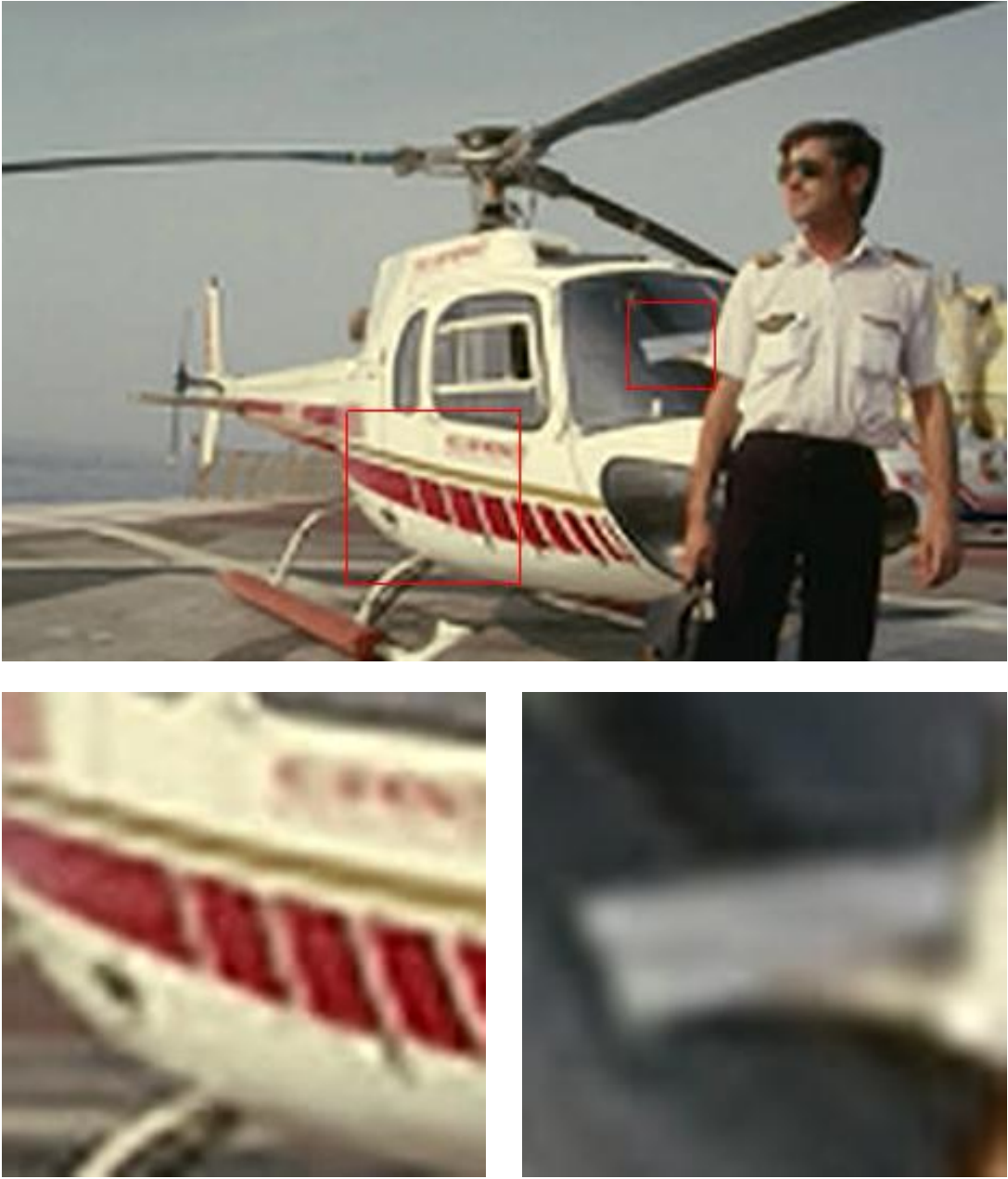}
}
\subfloat[][ \centering SRCNN \cite{dong2014srcnn} \par ]  {
\includegraphics[width=0.37\columnwidth]{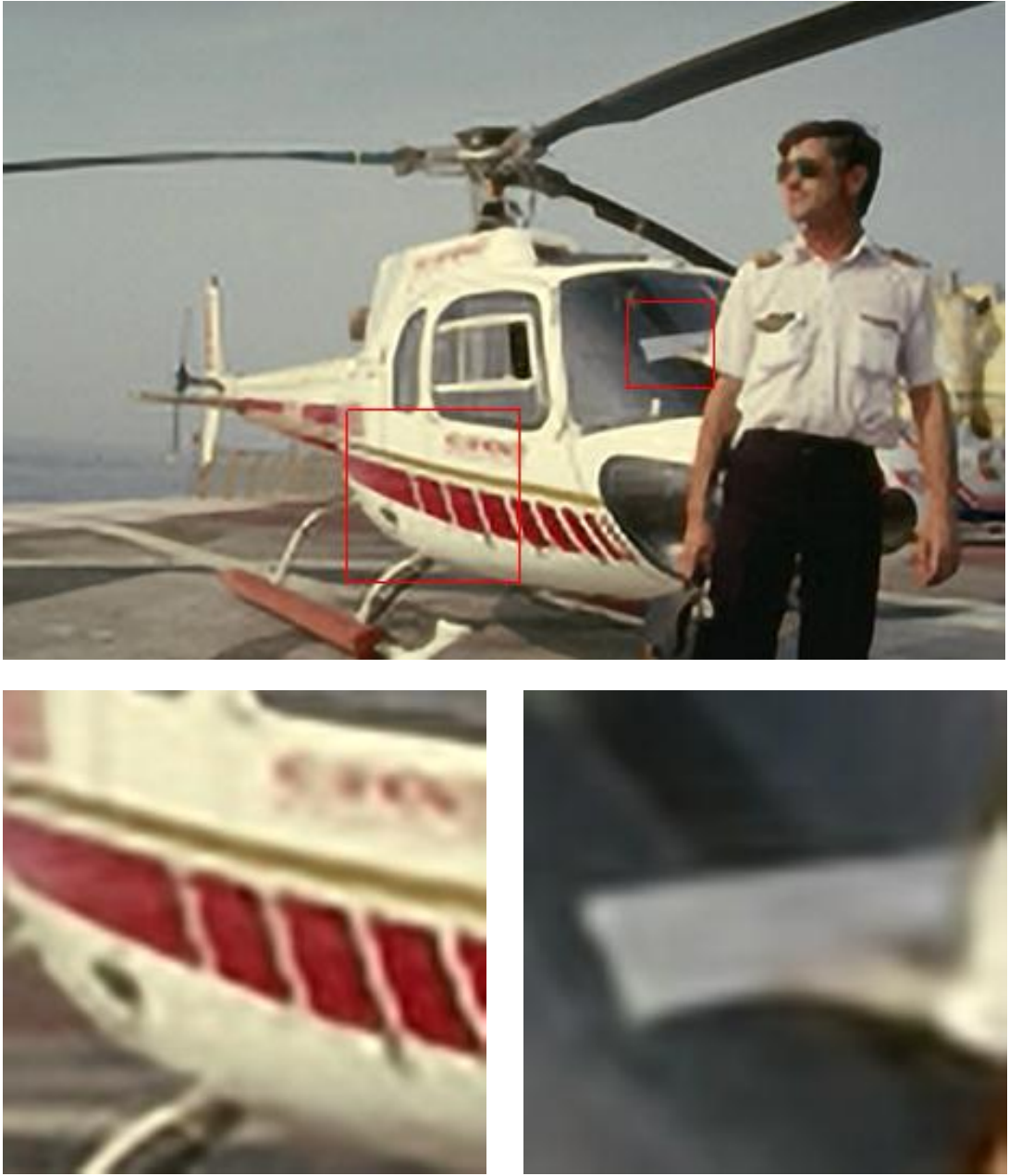}
}
\subfloat[][ \centering Ours\par ]  {
\includegraphics[width=0.37\columnwidth]{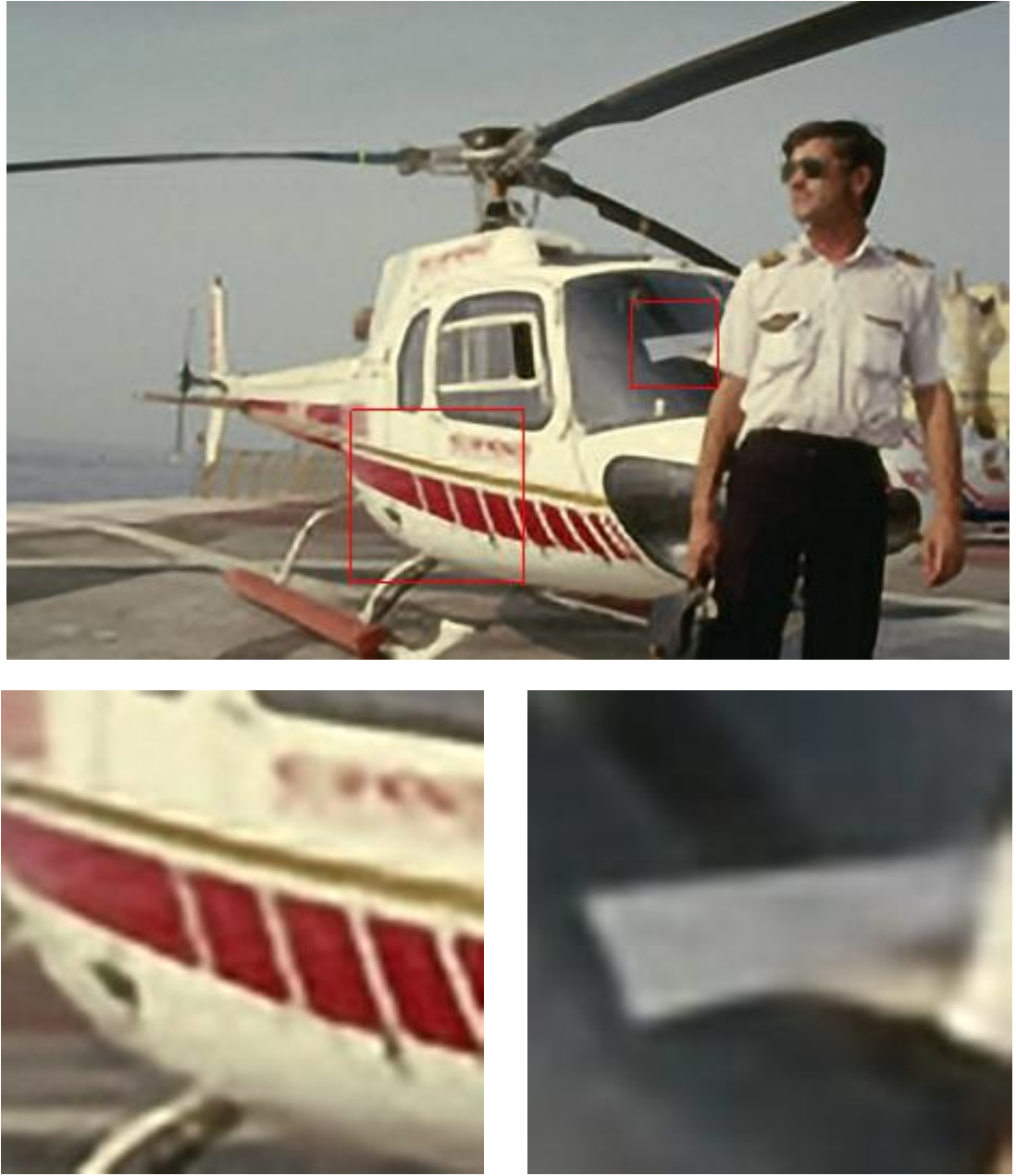}
}
\caption{The PSNR / SSIM indexes BSD200 `146074' image from \textit{Set5} }
\label{fig:set5-visualize}

\subfloat[][ \centering NELLE \cite{chang2004nelle}] {
\includegraphics[width=0.37\columnwidth]  {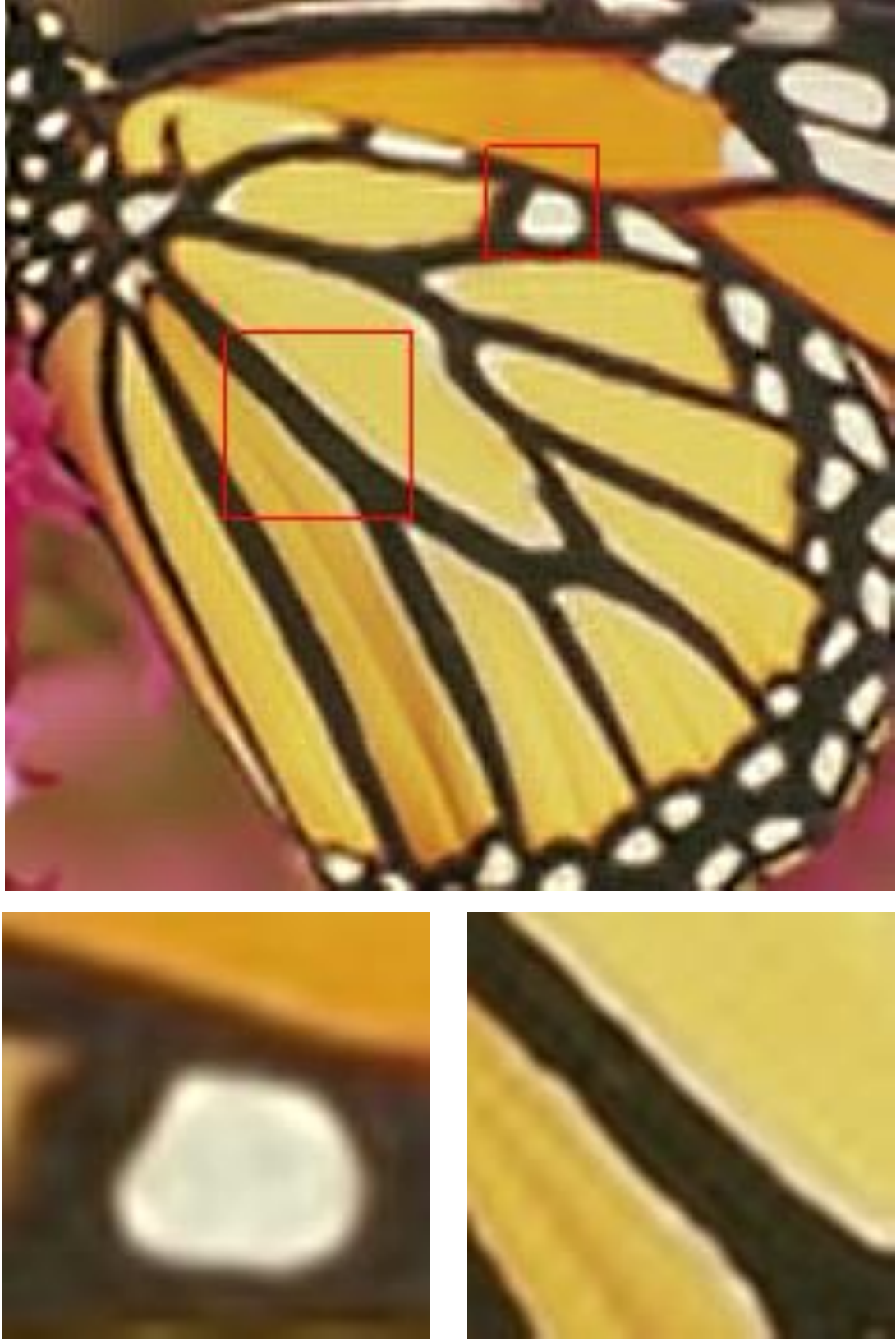}
}
\subfloat[][ \centering SCIP \cite{yang2010sc} ] {
\includegraphics[width=0.37\columnwidth]  {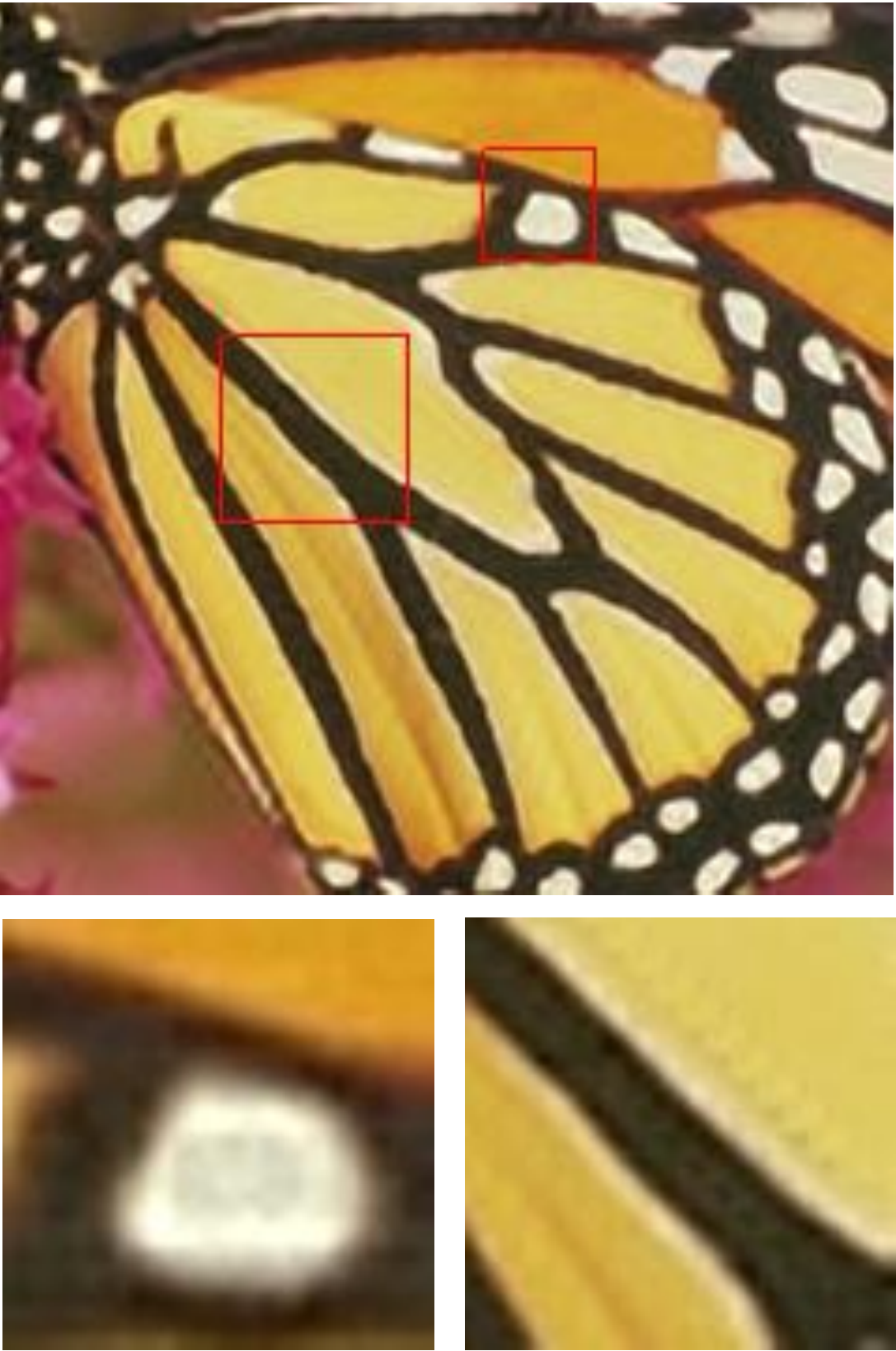}
}
\subfloat[][ \centering ANR \cite{timofte2013anr} ] {
\includegraphics[width=0.37\columnwidth]  {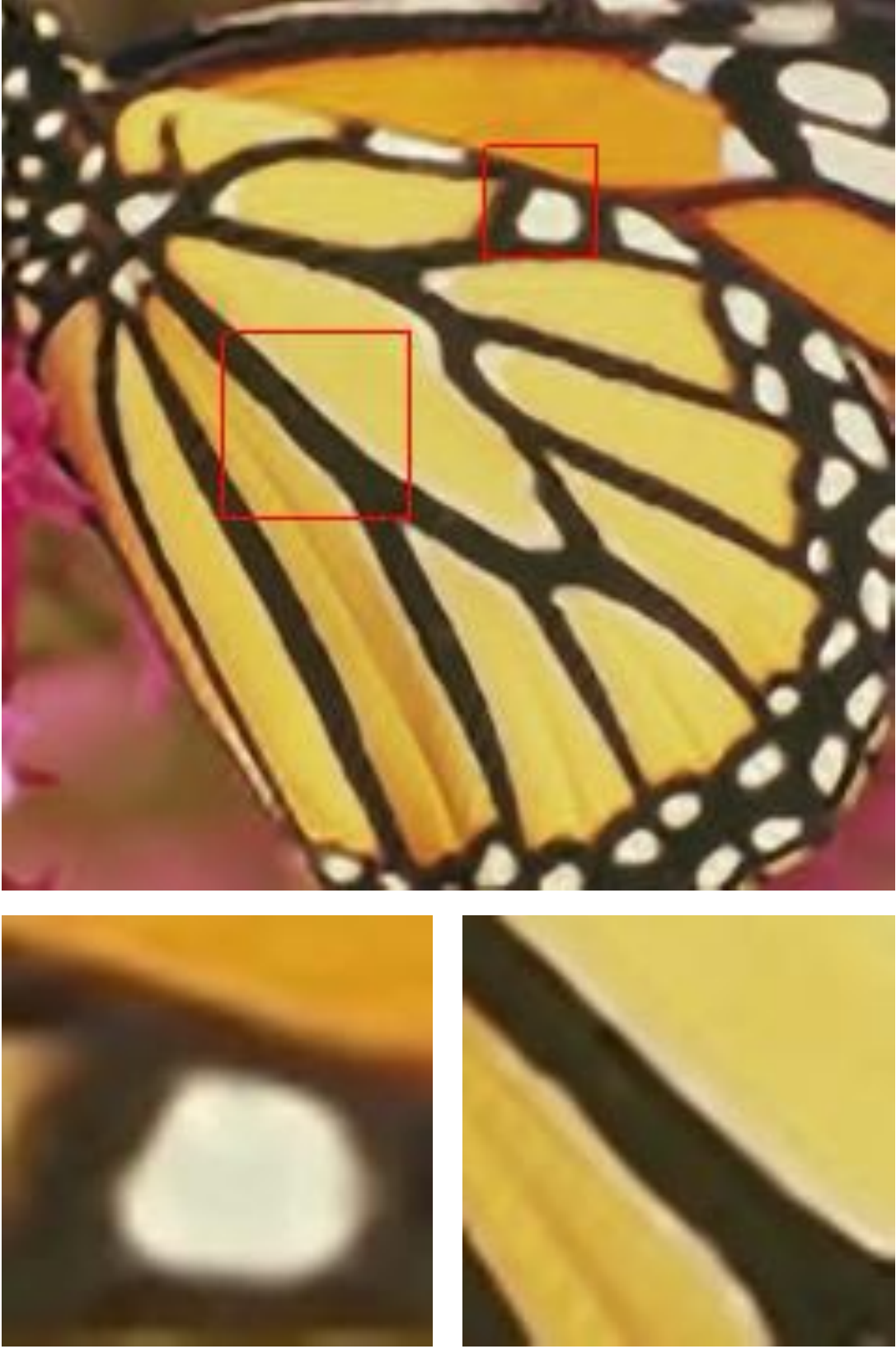}
}
\subfloat[][ \centering SRCNN \cite{dong2014srcnn} ] {
\includegraphics[width=0.37\columnwidth]  {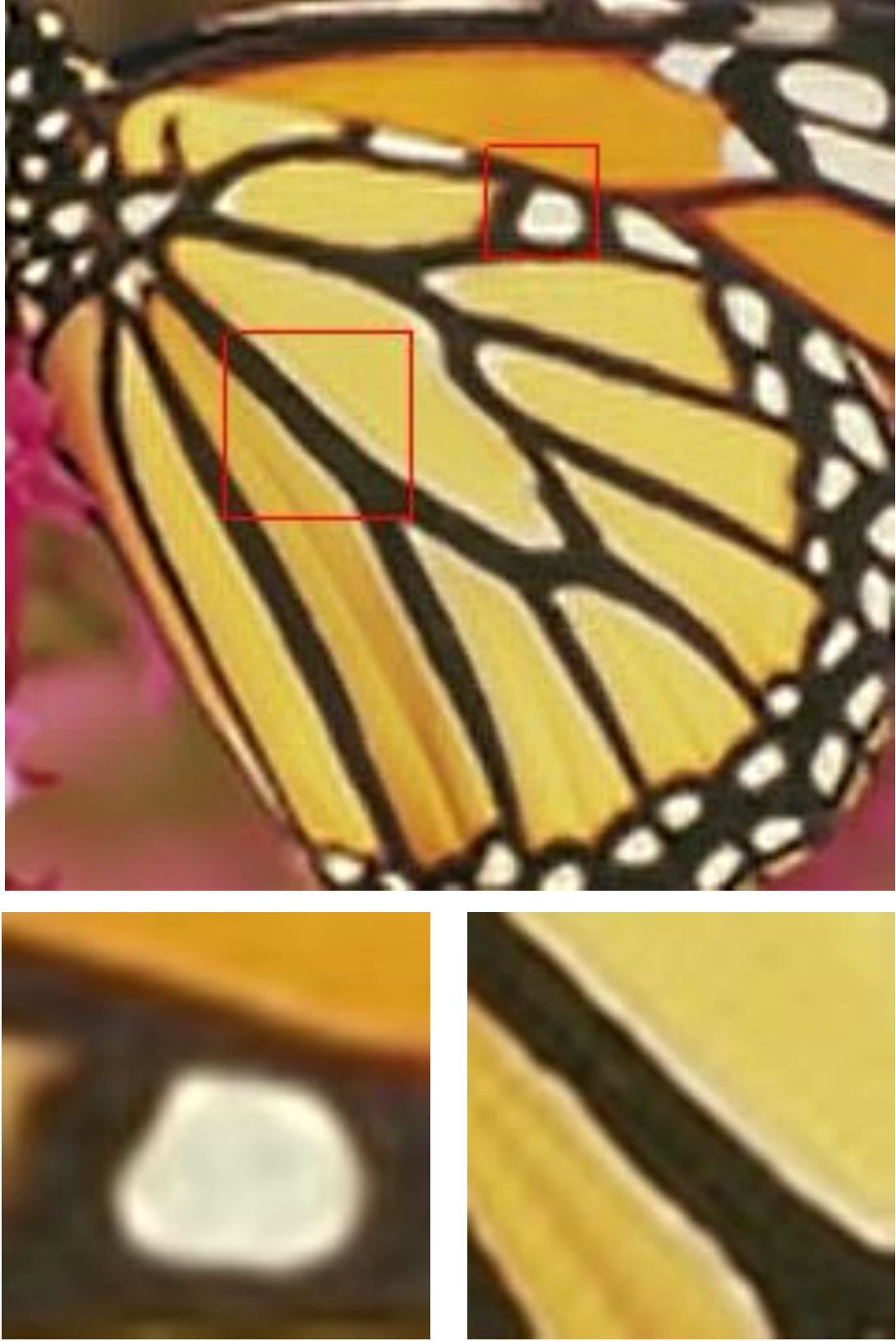}
}
\subfloat[][ \centering Ours ] {
\includegraphics[width=0.37\columnwidth]  {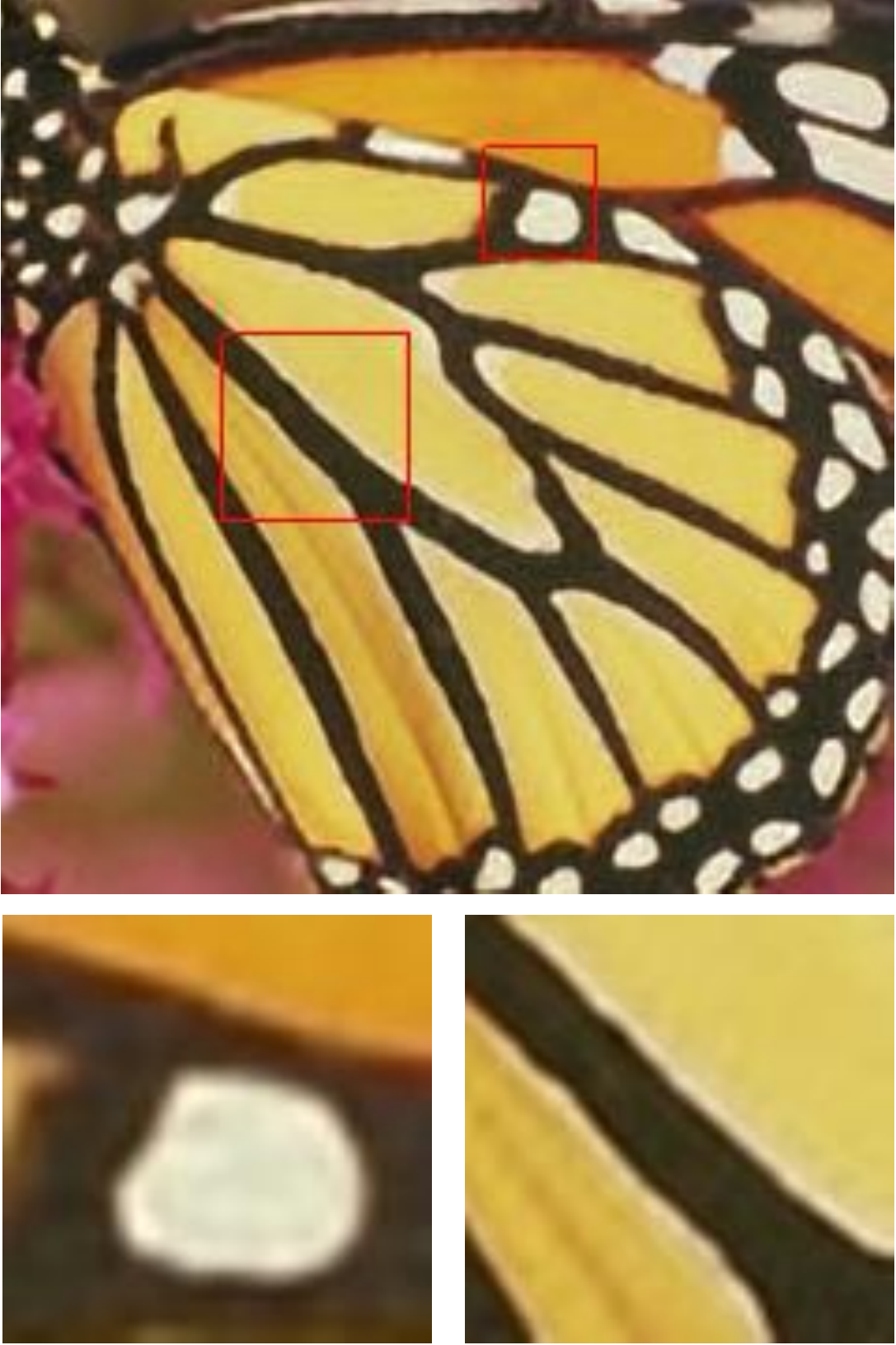}
}

\caption{'Butterfly' image from \emph{Set5}.}
\label{fig:internet-visualize}
\vspace{-15pt}
\end{figure*}

\subsection{Experiment Setting}
\indent \emph{\textbf{Datasets.}}~ To justify the effectiveness of our proposed model, we conduct extensive evaluations on three public benchmarks, i.e., the \textit{Set5} \cite{bevilacqua2012low}, \textit{Set14} \cite{zeyde2012single} and \textit{BSD500} \cite{amfm_pami2011} dataset. The \textit{BSD500} dataset consists of 500 images, we use its training / validation set (300 images) for training, the rest 200 images for testing (\textit{BSD200}). Besides, the \textit{Set5} and \textit{Set14} datasets are conducted by following the same experiment setting as other state-of-the-art methods \cite{dong2014srcnn, kim2010kk, timofte2013anr}.

\emph{\textbf{Implementation Details.}}~ For all above datasets, we first convert their images into YCbCr colorspace and only consider the luminance channel. Then we generate (32 x 32) sub-images from each HR images in the training set by stride of 12 pixel, and get the corresponding sub-images from boundary annotations of the dataset in the same way. We generate LR sub-images from HR sub-images by sub-sampling as \cite{dong2014srcnn}. To demonstrate that our model can handle image blur, we also introduce blurring into LR sub-images. As a result, we obtain the training set with 273600 triplets. Our proposed model is trained with the batch size $32$ and fixed learning rate $1e-12$ under each scaling factor of $\{\times2,\times3,\times4\}$.

\emph{\textbf{Evaluation Metric and Compared Methods.}}~
We adopt the widely used PSNR\footnote{Peak signal-to-noise ratio} and SSIM as our evaluation metrics. We compare our proposed joint component learning network of LSP and HSP (ours) with sparse coding and image prior (SCIP) \cite{kim2010kk}, neighbor embeding and locally linear embedding (NELLE) \cite{chang2004nelle}, anchored neighborhood regression (ANR) \cite{Yang13_ICCV_Fast} and a three layers CNN (SRCNN) \cite{dong2014srcnn}.

\begin{figure} [!ht]\centering
\subfloat[PSNR curves] {
\includegraphics[width=0.49\columnwidth]{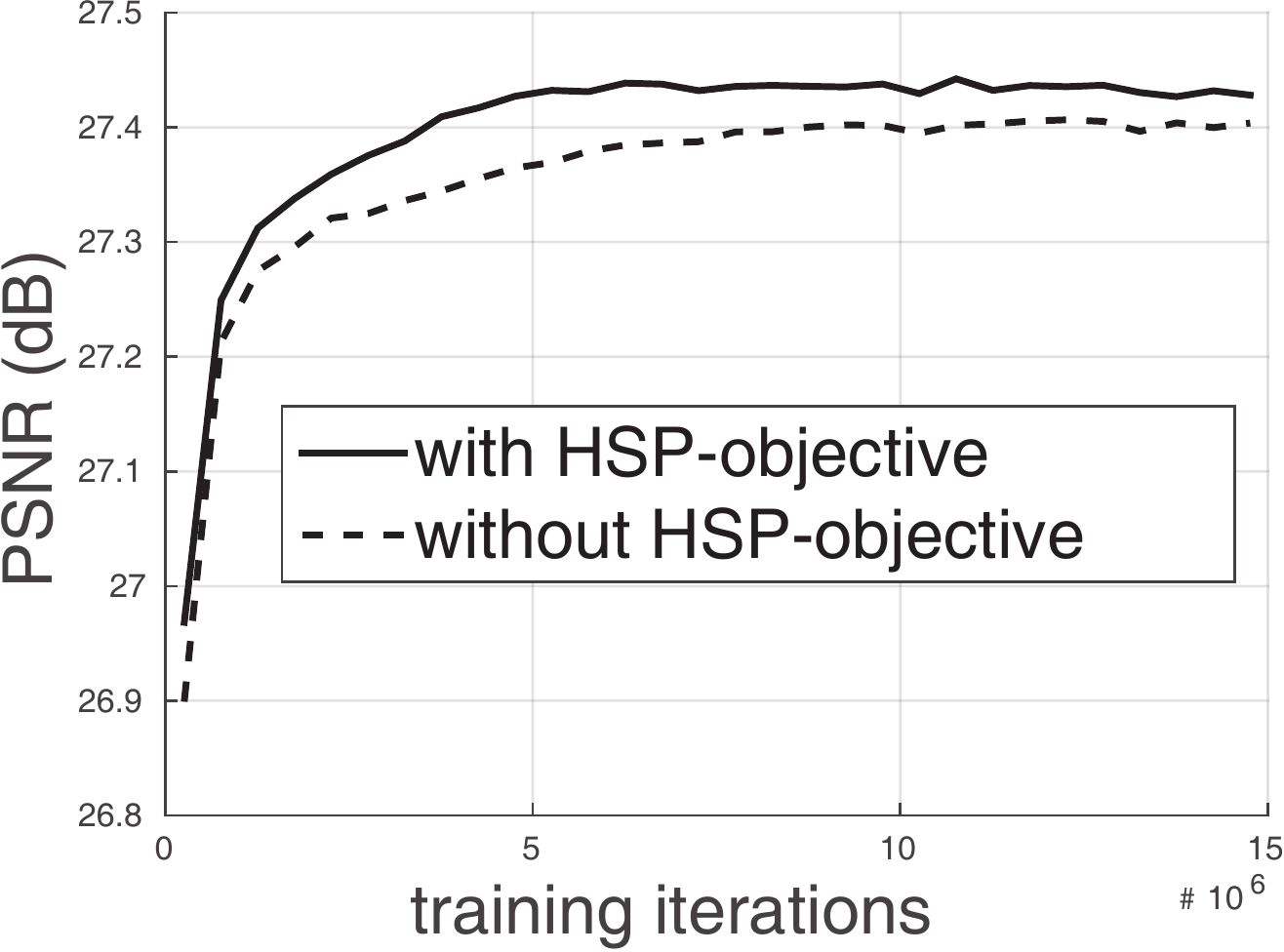}
}
\subfloat[SSIM curves] {
\includegraphics[width=0.49\columnwidth]{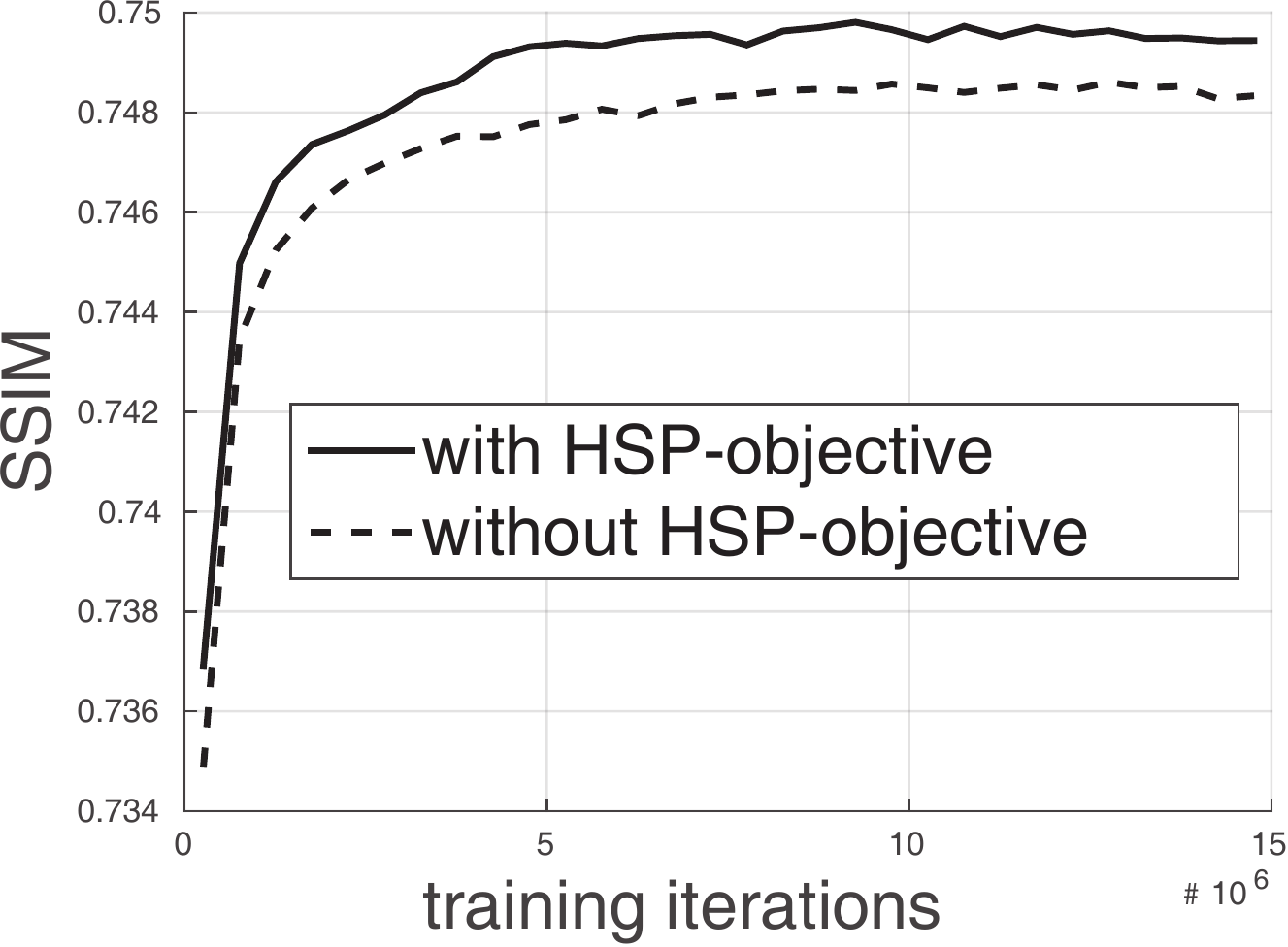}
}
\caption{The PSNR / SSIM curves generated by models trained with and without HSP-objective.}
\label{fig:comparison-on-HSP-and-noHSP}
\vspace{-15pt}
\end{figure}

\subsection{Empirical Results and Analysis}
Table \ref{table:PSNR/SSIM-on-different-method} indicates that our proposed model surpasses the former state-of-the-art methods both in the PSNR and SSIM indexes across all the scaling factors. From the results of PSNR metric, one can see that our model outperforms the compared methods by a large margin on \textit{Set5} (factor 2) and \textit{Set14}. A similar tend can also be observed for the SSIM evaluation metrics. To be specific, Table \ref{table:PSNR/SSIM-on-different-method} has shown the average gains, achieved by our model, are 0.47dB, 0.23dB, 0.2dB higher than second best method SRCNN \cite{dong2014srcnn}, which is also a deep learning method.

Figure \ref{fig:set5-visualize} and \ref{fig:internet-visualize} visualize some promising examples. We interpolate the Cb and Cr chrominance channels by the bicubic method to generate color images for better views. To clearly shown up the difference, we choose two patches from each group and attach them below. With the help of LSP, we can obtain multiple edge-preserving kernels. By means of these novel kernels, we achieve one better structure interpolation image. Compared to other methods, it gives rise to our results have sharper and clearer boundaries. We suggest the reader zooming in the figures to find more details.

To justify the contribution of the proposed HSP, we train our model with/without HSP-objective and plot the PSNR and SSIM curves. In Figure \ref{fig:comparison-on-HSP-and-noHSP}, it seems that the HSP-objective not only accelerates the convergence, but also provides a better initialization, which helps the network to converge at a better local optimal.


\section{Conclusion and Future Work}
In this work, we propose a novel structure preserving image super-resolution approach from both local and holistic perspectives. Extensive experiments demonstrate that our model not only achieve state-of-the-art performance on popular evaluation metrics, but also have a better visual quality. There are several directions in which we intend to extend this work. First, we only consider to estimate the pixel-place in the local gradient, we will explore more possibility in different elements and situations in the follow-up work. Second, we plan to extend our model for higher level vision tasks such as face hallucination.

\section*{Acknowledgements}
 We would like to thank Liliang Zhang for his assistance in the experiments.This work was supported in part by Guangdong Natural Science Foundation under Grant S2013050014548 and 2014A030313201, in part by Special Program for Applied Research on Super Computation of the NSFC-Guangdong Joint Fund (the second phase), and in part by Fundamental Research Funds for the Central Universities.


\bibliographystyle{IEEEbib}
\bibliography{icme2016template}

\end{document}